\documentclass[10pt,twocolumn,letterpaper]{article}

\usepackage[pagenumbers]{cvpr} 

\usepackage{graphicx}
\usepackage{amsmath}
\usepackage{amssymb}
\usepackage{booktabs}
\usepackage{algorithm}
\usepackage{algpseudocode}
\usepackage[accsupp]{axessibility}
\usepackage[pagebackref,breaklinks,colorlinks]{hyperref}

\usepackage[capitalize]{cleveref}
\crefname{section}{Sec.}{Secs.}
\Crefname{section}{Section}{Sections}
\Crefname{table}{Table}{Tables}
\crefname{table}{Tab.}{Tabs.}

\def\cvprPaperID{*****} 
\def\confName{CVPR}
\def\confYear{2022}

\begin{document}

\title{SwinLSTM: Improving Spatiotemporal Prediction Accuracy using Swin Transformer and LSTM}

\author{
	Song Tang\thanks{~Equal contribution. $^{\dag}$Corresponding author.}\hspace{1.2mm}\quad Chuang Li$^{*}$\quad Pu Zhang\quad RongNian Tang$^{\dag}$\\
	Hainan University\\
	\texttt{\{songtang,lc,zhangpu,rn.tang\}@hainanu.edu.cn} \\
}

\maketitle

\begin{abstract}
   Integrating CNNs and RNNs to capture spatiotemporal dependencies is a prevalent strategy for spatiotemporal prediction tasks. However, the property of CNNs to learn local spatial information decreases their efficiency in capturing spatiotemporal dependencies, thereby limiting their prediction accuracy. In this paper, we propose a new recurrent cell, SwinLSTM, which integrates Swin Transformer blocks and the simplified LSTM, an extension that replaces the convolutional structure in ConvLSTM with the self-attention mechanism. Furthermore, we construct a network with SwinLSTM cell as the core for spatiotemporal prediction. Without using unique tricks, SwinLSTM outperforms state-of-the-art methods on  Moving MNIST, Human3.6m, TaxiBJ, and KTH datasets. In particular, it exhibits a significant improvement in prediction accuracy compared to ConvLSTM. Our competitive experimental results demonstrate that learning global spatial dependencies is more advantageous for models to capture spatiotemporal dependencies. We hope that SwinLSTM can serve as a solid baseline to promote the advancement of spatiotemporal prediction accuracy. The codes are publicly available at~\href{https://github.com/SongTang-x/SwinLSTM}{https://github.com/SongTang-x/SwinLSTM}.
\end{abstract}

\section{Introduction}
\label{sec:intro}

\begin{figure}[tp]
	\centering
	\resizebox{\hsize}{!}{
		\includegraphics{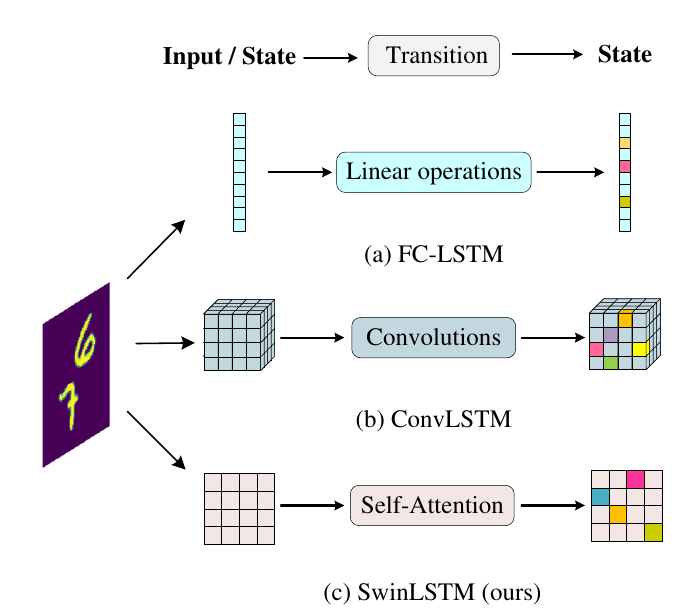}
	}
	\caption{Comparison of input-to-state and state-to-state transitions in three different recurrent cells. (a) FC-LSTM utilizes linear operations to process 1D vectors. (b) ConvLSTM employs 2D convolutions to process 3D tensors. (c) The proposed SwinLSTM leverages self-attention to process 2D matrices.}
	\label{fig:comparison}
\end{figure}
Spatiotemporal prediction has received increasing attention in recent years due to it can benefit many practical applications, e.g., precipitation forecasting~\cite{shi2015convolutional,shi2017deep,wang2017predrnn}, autonomous driving~\cite{bhattacharyya2018long,kwon2019predicting}, and traffic flow prediction~\cite{zhang2017deep,xu2018predcnn}. However, the complex physical dynamics and chaotic properties of spatiotemporal predictive learning make it challenging for purely data-driven deep learning methods to make accurate predictions. Existing methods~\cite{shi2015convolutional,wang2017predrnn,wang2018predrnn++,wang2019memory,lee2021video,wang2018eidetic,chang2021mau,guen2020disentangling,yu2020efficient} integrate CNNs and RNNs to learn spatiotemporal dependencies in spatiotemporal data to improve prediction accuracy. To capture the spatial and temporal dependencies simultaneously, ConvLSTM~\cite{shi2015convolutional} extends the fully connected LSTM (FC-LSTM)~\cite{srivastava2015unsupervised} by replacing linear operations with convolutional operations. 
Subsequently, several variants of ConvLSTM are proposed. PredRNN~\cite{wang2017predrnn} and MIM~\cite{wang2019memory} modify the internal structure of the LSTM unit. E3D-LSTM~\cite{wang2018eidetic} integrates 3D-Convs into LSTMs. PhyDNet~\cite{guen2020disentangling} leverages a CNN-based module to disentangle physical dynamics. Admittedly, these methods achieve impressive results on spatiotemporal prediction tasks. However, convolution operators focus on capturing local features and relations and are inefficient for modeling global spatial information~\cite{chen20182}. Although the receptive field can be enlarged by stacking convolution layers, the effective receptive field only reaches a fraction of the theoretical receptive field in practice~\cite{luo2016understanding}. Therefore, it can be inferred that the CNN-based models may prove to be ineffective in capturing spatiotemporal dependencies, owing to the inherent locality of the CNN architecture, leading to restricted accuracy in predictions.

Recently, the Vision Transformer (ViT)~\cite{dosovitskiy2020image} introduced the Transformer~\cite{vaswani2017attention} that directly models long-range dependencies into the vision domain. ViT applies a standard Transformer for image classification and attains excellent results with sufficient data. Its outstanding achievement has attracted more researchers to apply Transformer to computer vision. Subsequently, some variants~\cite{touvron2021training,wu2022p2t,zheng2021rethinking,liu2021swin} of ViT emerged, with tremendous success on different vision tasks. Notably, the Swin Transformer~\cite{liu2021swin} has exhibited exceptional performance across diverse visual tasks, owing to its unique utilization of local attention and shift window mechanism.

\begin{figure*}[htp]
	\centering
	\resizebox{\textwidth}{!}{
	\includegraphics{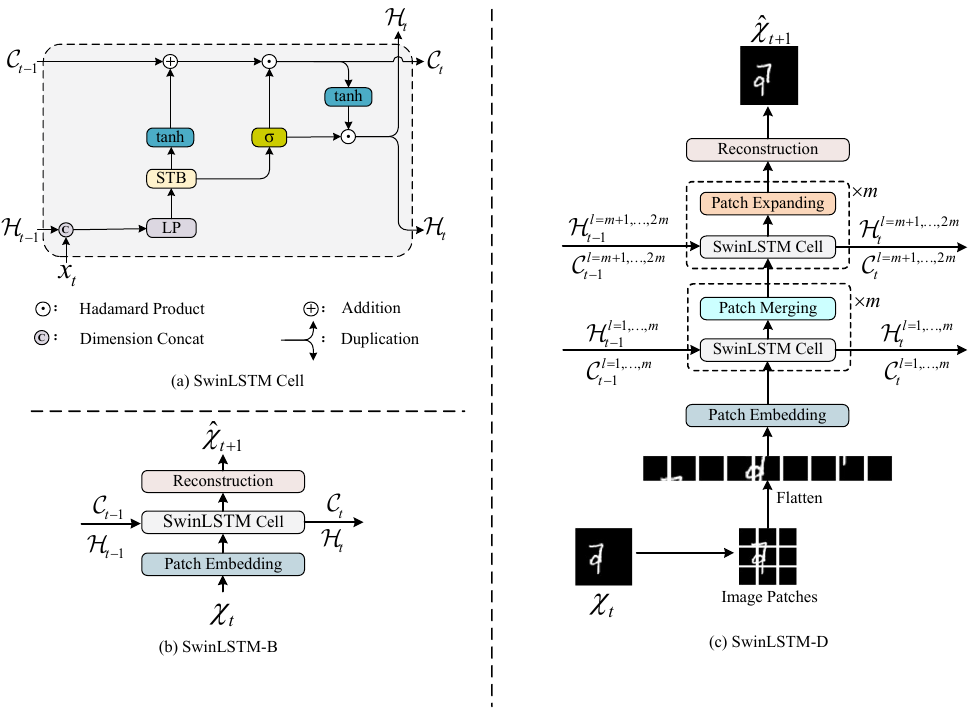}
	}
	\caption{(a): The detailed structure of the proposed recurrent cell: SwinLSTM. $\textbf{STB}$ and $\textbf{LP}$ denote Swin Transformer blocks and Linear Projection. (b): The architecture of the base model with a single SwinLSTM cell. (c): The architecture of the deep model with multiple SwinLSTM cells. }
	\label{fig:architecture}
\end{figure*}

Inspired by this, we propose SwinLSTM, a new recurrent cell. Specifically, we integrate Swin Transformer blocks and a simplified LSTM module to extract spatiotemporal representations. In addition, we construct a predictive network with SwinLSTM as the core to capture spatial and temporal dependencies for spatiotemporal prediction tasks. As illustrated in Figure~\ref{fig:architecture} (c), we first split an input image at the current time step into a sequence of image patches. Subsequently, the flattened image patches are fed to the patch embedding layer. Then, the SwinLSTM layer receives the transformed patches or the hidden states transformed by the previous layer (Patch Merging or Patch Expanding), and the cell and hidden states of the previous time step to extract the spatiotemporal representations. Finally, the reconstruction layer decodes the spatiotemporal representations to generate the next frame.

The contributions of this paper can be summarized as follows:
\begin{itemize}
	\item We propose a new recurrent cell, named SwinLSTM (section~\ref{sec:SwinLSTM}), which is able to efficiently extract spatiotemporal representations.
	
	\item We introduce a new architecture (section~\ref{sec:Architecture}) for spatiotemporal prediction tasks, which can efficiently model spatial and temporal dependencies.
	
	\item We evaluate the effectiveness of the proposed model on Moving MNIST, TaxiBJ, Human3.6m, and KTH. Experimental results show that SwinLSTM achieves excellent performance on four datasets.
\end{itemize}

\section{Related Work}
\paragraph{CNN-Based Models}  Prior models integrating CNNs and RNNs employ various strategies to better capture spatiotemporal dependencies to improve prediction accuracy. ConvLSTM~\cite{shi2015convolutional} extends FC-LSTM~\cite{srivastava2015unsupervised} by replacing fully connected operations with convolutional operations to learn spatiotemporal dependencies. PredRNN~\cite{wang2017predrnn} proposes the Spatiotemporal LSTM (ST-LSTM) module, which simultaneously models spatiotemporal information by transferring hidden states in horizontal and vertical directions. PredRNN++~\cite{wang2018predrnn++} designs a Gradient Highway unit to solve the vanishing gradient problem in PredRNN. E3D-LSTM~\cite{wang2018eidetic} replaces the 2D convolution in the ST-LSTM module with 3D convolution, allowing the ST-LSTM module to remember more previous information and achieve better prediction performance. MIM~\cite{wang2019memory} replaces the forget gate in the ST-LSTM module with two recurrent units to solve the non-stationary information learning in prediction. CrevNet~\cite{yu2020efficient} proposes a CNN-based reversible network to learn complex spatiotemporal dependencies. PhyDNet~\cite{guen2020disentangling} introduces physical knowledge into a CNN-based model to improve prediction quality. The above models~\cite{shi2015convolutional,wang2017predrnn,wang2018predrnn++,wang2018eidetic,wang2019memory,yu2020efficient,guen2020disentangling} improve their ability to capture spatiotemporal dependencies from different perspectives and achieve excellent results. However, despite their widespread usage, convolutional methods have inherent limitations in capturing spatiotemporal dependencies due to their local nature. In order to overcome this challenge, we propose the adoption of a global modeling approach (specifically, Swin Transformer blocks) for learning spatial dependencies, thereby enhancing the model's capacity to capture spatiotemporal dependencies and improve prediction performance. 
\paragraph{Vision Transformers}  The widespread use of Transformer~\cite{vaswani2017attention} in the field of natural language processing (NLP) led researchers to introduce it to the vision domain. ViT~\cite{dosovitskiy2020image} pioneered the direct application of the Transformer architecture to image classification and achieved great success. However, its excellent performance is based on training on large datasets, which leads to its unsatisfactory performance on smaller datasets. To solve this problem, DeiT~\cite{touvron2021training} proposes several training strategies to allow ViT to perform well on the smaller ImageNet-1K~\cite{deng2009imagenet} dataset. Subsequently, some variants~\cite{vaswani2021scaling,zhang2021multi,xie2021segformer,zhao2021point} of ViT achieved impressive results on various vision tasks. In particular, Swin Transformer~\cite{liu2021swin} achieves outstanding performance on image classification, semantic segmentation, and object detection tasks due to its shifted window mechanism and hierarchical design. In this paper, we try to integrate the Swin Transformer blocks and the simplified LSTM to form a SwinLSTM recurrent cell, and use it as the core to build a model to capture temporal and spatial dependencies to perform spatiotemporal prediction tasks.

\section{Method}
\subsection{Overall Architecture}
\label{sec:Architecture}
The overall predictive architecture is depicted in Figure~\ref{fig:architecture} (b and c). We build our base model and a deeper model, called SwinLSTM-B and SwinLSTM-D, respectively.First, an image at the time step $t$ is split into non-overlapping patches. The patch size is $P^{2}$, and $P$ equals 2 or 4 in our implementation. Therefore, the feature dimension of each patch is $C \cdot P^{2}$ ($C$ denotes the number of channels). Subsequently, the image patches are flattened and fed to the patch embedding layer, which linearly maps the original features of the patches to an arbitrary dimension. Then, for SwinLSTM-B, the SwinLSTM layer receives the transformed image patches,  the hidden state $\mathcal{H}_{t-1}$, and cell state $\mathcal{C}_{t-1}$ to generate the hidden state $\mathcal{H}_{t}$ and cell state $\mathcal{C}_{t}$, where the $\mathcal{H}_{t}$ is duplicated into two copies, one for the reconstruction layer, the other together with $\mathcal{C}_{t}$ for the SwinLSTM layer at the next time step. For SwinLSTM-D, we increase the number of SwinLSTM Cells, as well as add Patch Merging and Patch Expanding~\cite{cao2023swin} layers, where the former is utilized for downsampling and the latter for upsampling. $\mathcal{H}_{t}^{l=m}$ and $\mathcal{C}_{t}^{l=m}$ refer to the hidden state and cell state at time step $t$ from layer $m$. Finally, the reconstruction layer maps the hidden state $\mathcal{H}_{t}$ from the SwinLSTM layer to the input size to get the predicted frame for the next time step.
\begin{figure}
	\centering
	\includegraphics{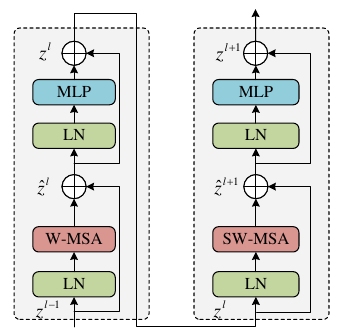}
	\caption{Swin Transformer Block.}
	\label{fig:swin}
\end{figure}

\begin{algorithm}
	\caption{Training pipeline} \label{alg1}
	\begin{algorithmic}[1]
		\State {\bf Require}:  $\alpha$, the learning rate. $m$, the batch size. $n$, the number of training epochs. $S$, the number of frames in the input (target) sequence.
		\State Initialize parameters of network ($\theta$). 
		\For{$i=0,1,..., n\text{-1}$}
		\For{number of batches}
		\State Sample ${\{[X_{0:S\text{-}1};Y_{0:S\text{-}1}]^{(j)}\}}_{j=1}^{m}$ a batch from the training set.
		\State {{\textbf{Phase 1}: Warm-up  Phase}}
		\State Initialize  $\mathcal{H}_{0}$, $\mathcal{C}_{0}$
		\For{$t=0,1,..., S\text{-}2$}
		\State Get $\hat{X}_{t+1}, \mathcal{H}_{t+1}, \mathcal{C}_{t+1}=P(X_{t}, \mathcal{H}_{t}, \mathcal{C}_{t})$
		\EndFor
		\State {{\textbf{Phase 2}: Prediction  Phase}}
		\State $\hat{I}_{0} \leftarrow {X}_{S\text{-}1}$, $\mathcal{H}_{0} \leftarrow \mathcal{H}_{S\text{-}1}$, $\mathcal{C}_{0} \leftarrow \mathcal{C}_{S\text{-}1}$
		\For{$t=0,1,..., S\text{-}1$}
		\State Get $\hat{I}_{t+1}, \mathcal{H}_{t+1}, \mathcal{C}_{t+1}=P(\hat{I}_{t}, \mathcal{H}_{t}, \mathcal{C}_{t})$
		\EndFor
		\State $\mathcal{L} \leftarrow \mathcal{L}^{\tt p}([\hat{X}_{1:S\text{-}1};\hat{I}_{1:S}], [{X}_{1:S\text{-}1};{Y}_{0:S\text{-}1}])$
		\State Update $\theta \leftarrow \theta - \alpha \nabla_{\theta} \mathcal{L}$
		\EndFor
		\EndFor
	\end{algorithmic}
\end{algorithm}
Both SwinLSTM-B and SwinLSTM-D were trained using the same training approach. The specific training process for SwinLSTM-B is presented in Algorithm ~\ref{alg1}. During the \textbf{Warm-up} phase, we take the frames of the input sequence as input to the model. However, in the \textbf{Prediction} phase, we use the output of the model at the previous time step as the input to the model at the current time step. In particular, we concatenate the outputs of the \textbf{Warm-up} and \textbf{Prediction} phases to compute the loss. When the prediction loss function $\mathcal{L}^{\tt p}$ is $L_2$, its formula is shown in eq.~\ref{eq:1}.

\begin{equation}
	\centering
	\begin{split}
		\mathcal{L}^{\tt p}=&\|[\hat{X}_{1:S\text{-}1};\hat{I}_{1:S}]-[{X}_{1:S\text{-}1};{Y}_{0:S\text{-}1}]\|^2_2
		\label{eq:1}
	\end{split}
\end{equation}
When the prediction loss function $\mathcal{L}^{\tt p}$ is $L_1+L_2$, its formula is shown in eq.~\ref{eq:2}.
\begin{equation}
	\begin{split}
		\mathcal{L}^{\tt p}=&\|[\hat{X}_{1:S\text{-}1};\hat{I}_{1:S}]-[{X}_{1:S\text{-}1};{Y}_{0:S\text{-}1}]\|_1\\
		                &+\|[\hat{X}_{1:S\text{-}1};\hat{I}_{1:S}]-[{X}_{1:S\text{-}1};{Y}_{0:S\text{-}1}]\|^2_2
		\label{eq:2}
	\end{split}
\end{equation}
In eq.~\ref{eq:1} and eq.~\ref{eq:2}, $\hat{X}_{1:S\text{-}1}$ represents S-1  predicted frames in the \textbf{Warm-up} phase, $\hat{I}_{1:S}$ denotes S predicted frames in the \textbf{Prediction} phase, and ${X}_{1:S\text{-}1}$ and ${Y}_{0:S\text{-}1}$ represent S-1 input frames and S   ground truth future frames, respectively.

\subsection{SwinLSTM Module}
\label{sec:SwinLSTM}
ConvLSTM~\cite{shi2015convolutional} introduces the convolution operators into input-to-state and state-to-state transitions to overcome the shortcomings of FC-LSTM~\cite{srivastava2015unsupervised} in processing spatiotemporal data. The key equations of ConvLSTM are as follows:
\begin{equation}
	\begin{split} 
		&i_{t} =\sigma\left(W_{x i} * \mathcal{X}_{t}+W_{h i} * \mathcal{H}_{t-1}+b_{i}\right) \\
		&f_{t} =\sigma\left(W_{x f} * \mathcal{X}_{t}+W_{h f} * \mathcal{H}_{t-1}+b_{f}\right) \\
		&o_{t} =\sigma\left(W_{x o} * \mathcal{X}_{t}+W_{h o} * \mathcal{H}_{t-1}+b_{o}\right) \\
		&\mathcal{C}_{t} =f_{t} \circ \mathcal{C}_{t-1}+i_{t} \circ \tanh \left(W_{x c} * {\mathcal{X}}_{t}+W_{h c} * {\mathcal{H}}_{t-1}+b_{c}\right) \\
		&\mathcal{H}_{t} =o_{t} \circ \tanh \left(\mathcal{C}_{t}\right)
	\end{split}
	\label{eq:eq1}
\end{equation}
where ‘*’ denotes the convolution operator and ‘◦’ denotes the Hadamard product. $i_{t}$, $f_{t}$, $o_{t}$ are input,  forget, output gate. Different from convolution operators to extract local correlations, the self-attention mechanism captures global spatial dependencies by computing similarity scores across all positions. Thus, we remove all weights $W$ and biases $b$ in eq.~\ref{eq:eq1} to obtain eq.~\ref{eq:eq2}:
\begin{equation}
	\begin{split} 
		i_{t} &= f_{t} = o_{t} = \sigma\left(\mathcal{X}_{t}+ \mathcal{H}_{t-1}\right) \\
		\mathcal{C}_{t} &=f_{t} \circ \mathcal{C}_{t-1}+i_{t} \circ \tanh \left({\mathcal{X}}_{t}+ {\mathcal{H}}_{t-1}\right) \\
		\mathcal{H}_{t} &=o_{t} \circ \tanh \left(\mathcal{C}_{t}\right)
	\end{split}
	\label{eq:eq2}
\end{equation}
Obviously, $i_{t} = f_{t} = o_{t}$ in eq.~\ref{eq:eq2}. Therefore, we fuse the three gates: $i_{t}$, $f_{t}$, $o_{t}$ into one gate, named filter gate $\mathcal{F}_{t}$.

The detailed structure of the SwinLSTM  Module is presented in Figure~\ref{fig:architecture} (a).  In SwinLSTM, the information of cell states $\mathcal{C}_{t}$ and hidden states $\mathcal{H}_{t}$ is updated horizontally to capture long-term and short-term temporal dependencies. Meanwhile, the Swin Transformer blocks vertically learns global spatial dependencies. The key equations of SwinLSTM are shown in eq.~\ref{eq:eq3}, where STB denotes the Swin Transformer blocks and LP denotes the Linear Projection:
\begin{equation}
	\begin{split} 
		\mathcal{F}_{t} & =\sigma\left(\operatorname{STB}\left(\operatorname{LP}\left(\mathcal{X}_{t} ;  \mathcal{H}_{t-1}\right)\right)\right) \\
		\mathcal{C}_{t} & =\mathcal{F}_{t} \circ\left(\tanh \left(\operatorname{STB}\left(\operatorname{LP}\left(\mathcal{X}_{t} ;  \mathcal{H}_{t-1}\right)\right)\right)+\mathcal{C}_{t-1}\right) \\
		\mathcal{H}_{t} & =\mathcal{F}_{t} \circ \tanh \left(\mathcal{C}_{t}\right)
	\end{split}
	\label{eq:eq3}
\end{equation}
\subsection{Swin Transformer Block}
\label{sec:STB}
The global multi-head self-attention (MSA) mechanism in the Vision Transformer~\cite{dosovitskiy2020image} computes the relationship between a token and all other tokens. Such a computation leads to quadratic computational complexity related to the number of tokens, which is not friendly to dense prediction tasks. To alleviate the computational burden of global self-attention and improve the modeling power, Swin Transformer~\cite{liu2021swin} proposes window-based multi-head self-attention (W-MSA) and shifted-window-based multi-head self-attention (SW-MSA).
\begin{equation}
	\begin{split} 
		&\hat{\mathbf{z}}^{l} =\operatorname{W-MSA}\left(\mathrm{LN}\left(\mathbf{z}^{l-1}\right)\right)+\mathbf{z}^{l-1} \\
		&\mathbf{z}^{l} =\operatorname{MLP}\left(\operatorname{LN}\left(\hat{\mathbf{z}}^{l}\right)\right)+\hat{\mathbf{z}}^{l} \\
		&\hat{\mathbf{z}}^{l+1} =\operatorname{SW-MSA}\left(\mathrm{LN}\left(\mathbf{z}^{l}\right)\right)+\mathbf{z}^{l} \\
		&\mathbf{z}^{l+1} =\operatorname{MLP}\left(\operatorname{LN}\left(\hat{\mathbf{z}}^{l+1}\right)\right)+\hat{\mathbf{z}}^{l+1}
	\end{split}
	\label{eq:eq4}
\end{equation}
\begin{table*}[h]
	\centering
	$
	\begin{array}{cccccccc}
		\toprule \text { Dataset }& \text { Model }& \text { SwinLSTMs } & \text { STB } & \text { Patch size } & \text { Resolution } & \text { Train } & \text { Test }\\
		\midrule 
		\text {Moving MNIST} & \text {SwinLSTM-D}& \text {4}& \text {(2, 6, 6, 2)} & \text{2} & \text{(64, 64, 1)} & \text {10} \rightarrow \text {10} & \text {10} \rightarrow \text {10}\\
		\text {Human3.6m}& \text {SwinLSTM-B}& \text {1} &\text {12} & \text{2}  & \text{(128, 128, 3)} & \text {4} \rightarrow \text {4} & \text {4} \rightarrow \text {4} \\
		\text {KTH}& \text {SwinLSTM-B}& \text {1} & \text{6} & \text{4}  & \text{(128, 128, 1)} & \text {10}\rightarrow \text {10} & \text {10}\rightarrow \text {20}/ \text {40}\\
		\text {TaxiBJ}& \text {SwinLSTM-B}& \text {1} & \text {12} & \text{4}  & \text{(32, 32, 2)} & \text {4} \rightarrow \text {4}& \text {4}\rightarrow \text {4} \\
		\bottomrule
	\end{array}
	$
	
	\caption{Experimental setup. $\textbf{SwinLSTMs}$ denotes the number of the SwinLSTM Cells in predictive network. $\textbf{STB}$ denotes the number of the Swin Transformer blocks in SwinLSTM cell. $\textbf{Patch size}$ indicates the patch token size. $\textbf{Train}$ and $\textbf{Test}$ represent the number of input and output frames during training and testing.}
	\label{tab:para}
\end{table*}
As shown in Figure~\ref{fig:swin}, the first Swin Transformer block is based on W-MSA, consisting of two LayerNorm layers and a 2-layer MLP with GELU non-linearity. A LayerNorm layer and a residual connection are applied before and after each W-MSA module and MLP module, respectively. The second Swin Transformer block is the same as the first except that the W-MSA block in it is replaced by an SW-MSA block. The key equations of Swin Transformer blocks are shown in eq.~\ref{eq:eq4}.

\section{Experiments}
We evaluate our proposed model on four commonly used datasets: Moving MNIST~\cite{srivastava2015unsupervised}, TaxiBJ~\cite{zhang2017deep}, Human3.6m~\cite{ionescu2013human3}, and KTH~\cite{schuldt2004recognizing}, and present quantitative comparison results and visual examples for each dataset, illustrating the effectiveness and generalizability of our proposed neural network. In addition, we perform ablation studies (section~\ref{sec:Ablation}) and feature map visualization (section~\ref{Feature}).

\subsection{Implementations}
We use $L_{2}$ loss for the MovingMNIST dataset, $L_{1}+L_{2}$ loss for the Human3.6m,  the TaxiBJ, and the KTH datasets. The input and target frames of each dataset are normalized to the intensity of [0, 1]. Each model is optimized with the Adam optimizer~\cite{kingma2015adam}. We train our model on an Nvidia RTX A5000 GPU. More detailed parameters for each dataset are listed in Table~\ref{tab:para}.

\subsection{Evaluation Metrics}
We adopt the Mean Squared Error (MSE), the Mean Absolute Error (MAE), the Peak Signal to Noise Ratio (PSNR), and the Structural Similarity Index Measure (SSIM)~\cite{wang2004image} as metrics to evaluate the quality of predictions. All metrics are averaged over predicted frames. Lower MAE and MSE or higher SSIM and PSNR indicate better prediction accuracy.
\begin{table}
	\centering
	$
	\begin{array}{ccc}
		\toprule \text { Method }  & \text { MSE } \downarrow & \text { SSIM }\uparrow  \\
		\midrule 
		\text { ConvLSTM~\cite{shi2015convolutional} }  & \text{103.3} & \text{0.707} \\ 
		\text { DFN~\cite{jia2016dynamic} }  & \text{89.0} & \text{0.726} \\ 
		\text { FRNN~\cite{oliu2018folded} }  & \text{69.7} & \text{0.813} \\ 
		\text { VPN~\cite{kalchbrenner2017video} }  & \text{64.1} & \text{0.870} \\
		\text { PredRNN~\cite{wang2017predrnn} }  & \text{56.8} & \text{0.867} \\
		\text { CausalLSTM~\cite{wang2018predrnn++} }  & \text{46.5} & \text{0.898} \\
		\text { MIM~\cite{wang2019memory} } & \text{44.2} & \text{0.910}  \\
		\text { E3D-LSTM~\cite{wang2018eidetic} }  & \text{41.3} & \text{0.910} \\
		\text { LMC~\cite{lee2021video} }  & \text{41.5} & \text{0.924} \\
		\text { MAU~\cite{chang2021mau} }  & \text{27.6} & \text{0.937} \\
		\text { PhyDNet~\cite{guen2020disentangling}}  & \text{24.4} & \text{0.947}\\
		\text { CrevNet~\cite{yu2020efficient}}  & \text{22.3} & \text{0.949}\\
		\midrule
		\text {SwinLSTM } &\textbf{17.7} & \textbf{0.962} \\
		\bottomrule
	\end{array}
	$
	\caption{ Quantitative comparison of SwinLSTM and other methods on \textbf{Moving MNIST}. Each model observes 10 frames and predicts the subsequent 10 frames. Lower MSE and higher SSIM indicate better predictions.}
	\label{tab:mm}
\end{table}
\begin{table}
	\centering
	\resizebox{\hsize}{!}{
		$
		\begin{array}{cccc}
			\toprule \text { Method }& \text { MAE }/100\downarrow   & \text { MSE }/10\downarrow & \text{SSIM}\uparrow  \\
			\midrule 
			\text { FRNN~\cite{oliu2018folded} }  &\text {19.0} & \text {49.8} & \text {0.771} \\ 
			\text { ConvLSTM~\cite{shi2015convolutional} }  &\text {18.9} & \text {50.4} & \text {0.776} \\ 
			\text { PredRNN~\cite{wang2017predrnn} }&\text {18.9}  &  \text {48.4}& \text {0.781}  \\ 
			\text { CausalLSTM~\cite{wang2018predrnn++} }&\text {17.2}  & \text {45.8}  & \text {0.851} \\
			\text { MIM~\cite{wang2019memory} }&\text {17.8} & \text {42.9} & \text {0.790}   \\
			\text { E3D-LSTM~\cite{wang2018eidetic} }& \text {16.6} &  \text {46.4} & \text {0.869} \\
			\text { PhyDNet~\cite{guen2020disentangling} }&\text {16.2}  & \text {36.9} & \text {0.901} \\
			\midrule
			\text {SwinLSTM } &\textbf{11.9}  &\textbf{33.2}  &\textbf{0.913} \\
			\bottomrule
		\end{array}
		$
	}
	\caption{Quantitative evaluation of different models on the \textbf{Human3.6m} dataset.}
	\label{tab:human}
\end{table}
\begin{table}
	\centering
	\resizebox{\hsize}{!}{
		$
		\begin{array}{ccccc}
			\toprule & \multicolumn{2}{c}{\text { KTH }(\text {10}\rightarrow\text {20})} & \multicolumn{2}{c}{\text { KTH }(\text {10}\rightarrow\text {40})} \\
			\cline { 2 - 5 } \text { Method } & \text { SSIM }\uparrow & \text { PSNR }\uparrow & \text { SSIM }\uparrow & \text { PSNR }\uparrow \\
			\midrule
			\text { ConvLSTM~\cite{shi2015convolutional} } &\text {0.712} & \text {23.58} & \text {0.639} & \text {22.85} \\
			\text { SAVP~\cite{gao2019disentangling} } &\text {0.746} & \text {25.38} & \text {0.701} & \text {23.97} \\
			\text { FRNN~\cite{oliu2018folded} } & \text {0.771} & \text {26.12} &\text {0.678} & \text {23.77}\\
			\text { DFN~\cite{jia2016dynamic} } & \text {0.794} & \text {27.26}  & \text {0.652} & \text {23.01} \\
			\text { PredRNN~\cite{wang2017predrnn}  } & \text {0.839} & \text {27.55} & \text {0.703} & \text {24.16} \\
			\text { VarNet~\cite{jin2018varnet}  } & \text {0.843} & \text {28.48} & \text {0.739} & \text {25.37} \\
			\text { SVAP-VAE~\cite{lee2018stochastic}  } & \text {0.852} & \text {27.77} & \text {0.811} & \text {26.18} \\
			\text { PredRNN++~\cite{wang2018predrnn++}  } & \text {0.865} & \text {28.47} & \text {0.741} & \text {25.21} \\
			\text { E3d-LSTM~\cite{wang2018eidetic}  } & \text {0.879} & \text {29.31} & \text {0.810} & \text {27.24} \\
			\text { STMFANet~\cite{jin2020exploring}  } & \text {0.893} & \text {29.85} & \text {0.851} & \text {27.56} \\
			\midrule 
			\text {SwinLSTM } & \textbf{0.903} & \textbf{34.34} & \textbf{0.879} & \textbf{33.15} \\
			\bottomrule
		\end{array}
		$
	}	
	\caption{Quantitative evaluation on the \textbf{KTH} test set. We present the model observing 10 frames to predict 20 or 40 frames, and all metrics are averaged over the predicted frames. Higher SSIM and PSNR indicate better prediction quality.}
	\label{tab:KTH}
\end{table}

\begin{figure*}
	\centering
	\resizebox{\textwidth}{!}{
		\includegraphics{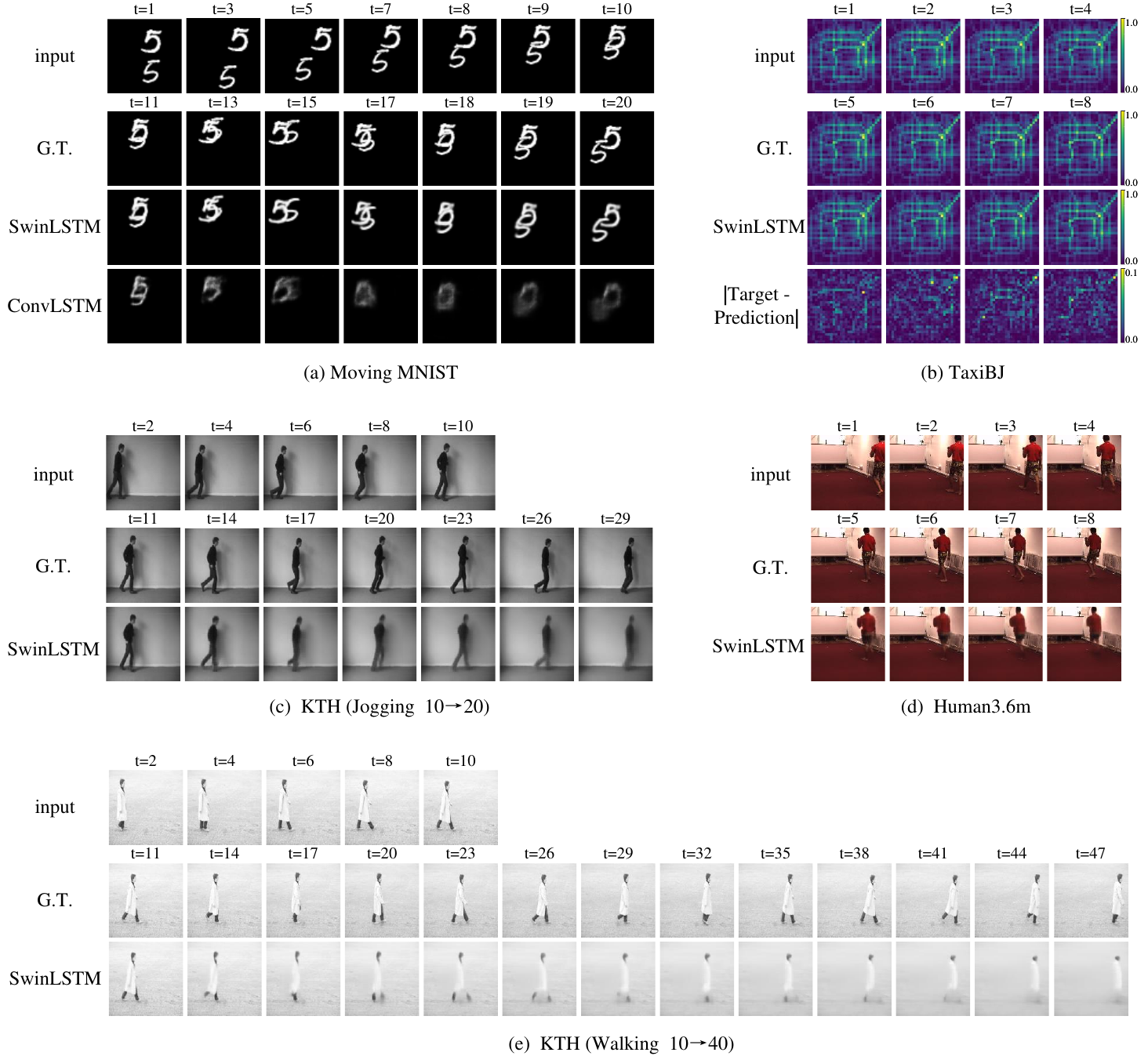}
	}
	\caption{Qualitative results of SwinLSTM on four datasets. The first line is the input, the second line is the ground truth, and the third line is the prediction of SwinLSTM. For Moving MNIST, we add the prediction results of ConvLSTM as the fourth line. For TaxiBJ, we add the absolute difference between the predictions and ground truths as the fourth line. For KTH action, (c) is the jogging action when the model predicts 20 frames based on the previous 10 frames, and (e) is the walking action when the model observes 10 frames and predicts the next 40 frames.}
	\label{fig:visualization}
\end{figure*}
\subsection{Datasets}
\noindent{\bf Moving MNIST.}  The moving MNIST dataset~\cite{srivastava2015unsupervised} is a widely used synthetic dataset. In our experiments, the method for generating moving MNIST sequences follows~\cite{srivastava2015unsupervised}, each sequence contains 20 frames, and the first 10 frames and the last 10 frames are used for input and target, respectively. The two handwritten digits in each frame are randomly sampled in the MNIST dataset~\cite{lecun1998mnist}, moving and bouncing around a $64\times64$ pixel black canva with a fixed speed and angle. We can get an infinite number of sequences by applying different speeds and angles to different handwritten digits. We adopt 10000 sequences for training and a fixed set containing 10000 sequences for testing.

\noindent{\bf Human3.6m.}  The human3.6m dataset~\cite{ionescu2013human3} contains 3.6 million different human poses and corresponding images. Following~\cite{wang2019memory}, we only use the 'walking' scenario. Subjects S1, S5, S6, S7, S8 are used for training, and S9, S11 are used for testing. The images in the dataset are resized from $1000\times1000\times3$ to $128\times128\times3$. We train the models to predict the next 4 RGB frames from 4 observations.

\noindent{\bf KTH.}  The KTH dataset~\cite{schuldt2004recognizing} contains 25 individuals performing 6 categories of human actions (walking,
jogging, running, boxing, hand waving, and hand clapping)  in 4 different scenarios. We follow the experimental setup in\cite{villegas2017decomposing}, resizing each image to $128\times128$ and using persons 1-16 for training and 17-25 for testing. The models predict 10 frames from 10 observations at training time and 20 or 40 frames at inference time.

\noindent{\bf TaxiBJ.}  TaxiBJ~\cite{zhang2017deep} contains complex real-world taxi trajectory data collected from taxicab GPS monitors in Beijing. Each frame in TaxiBJ is a $32\times32\times2$ heat map, where the last dimension represents the flow of traffic entering and leaving the same area. We follow the experimental setup in~\cite{zhang2017deep} and use the last four weeks of data for testing and the rest for training. We use 4 observations to predict the next 4 consecutive frames. We adopted per-frame MSE as metrics.

\subsection{Main results}
Table~\ref{tab:mm},~\ref{tab:human},~\ref{tab:KTH} and ~\ref{tab:taxi} present quantitative comparisons with previous state-of-the-art models on four datasets. On Moving MNIST, SwinLSTM outperforms previous CNN-based models and achieves the new state-of-the-art results. Specifically, compared to ConvLSTM~\cite{shi2015convolutional}, SwinLSTM reduces the MSE from 103.3 to 17.7 and increases the SSIM from 0.707 to 0.962. On both Human3.6m and KTH, SwinLSTM achieves the state-of-the-art, especially with huge gains of 4.49 (10$\rightarrow$20) and 5.59 (10$\rightarrow$40) in PSNR on KTH. In the case of TaxiBJ, our proposed SwinLSTM achieved a remarkable reduction in MSE for each predicted frame compared to previous models. Additionally, the inter-frame MSE difference was smaller than that of the compared models. The results indicate SwinLSTM's great potential for an efficient and generalizable technique for spatiotemporal prediction.

In Figure~\ref{fig:visualization}, we present qualitative results of SwinLSTM on all datasets. (a) presents a qualitative analysis of SwinLSTM and ConvLSTM on an extremely hard case of Moving MNIST. In the presented images, two digits are continuously intertwined, posing a challenge to the model's ability to accurately predict their motion trajectories and appearances. As demonstrated, both models effectively capture the trajectory of the digits. However, in terms of appearance, ConvLSTM's prediction results become increasingly blurred as the number of time steps increases, while SwinLSTM's predictions maintain high similarity throughout. SwinLSTM, as an extension of ConvLSTM, utilizes LSTM~\cite{hochreiter1997long} to capture temporal dependencies, and employs Swin Transformer blocks~\cite{liu2021swin} to model global spatial information. In contrast, ConvLSTM uses convolution to model local spatial information. The significant performance gap between the two models demonstrates that learning global spatial information can better assist the model in capturing spatiotemporal dependencies and thus improve prediction accuracy. 
 
To facilitate the observation of the similarity between the predicted results and the ground truth in (b), we visualize the absolute difference between them. For (c), (d), and (e), we observe that SwinLSTM makes accurate predictions of human actions and scenes, where (c) and (e) show visual examples of 2 different categories of actions predicted by SwinLSTM, Jogging and walking respectively. The model receives 10 frames during training and predicts 10 frames, whereas it predicts 20 or 40 frames during testing. Expanding the predicted number of frames during testing poses a significant challenge to the model, requiring it to effectively learn spatiotemporal dependencies from the spatiotemporal data. Our experiments on the KTH dataset demonstrate that SwinLSTM outperforms alternative models in capturing spatiotemporal dependencies.

\begin{table}
	\centering
	\resizebox{\hsize}{!}{
		$
		\begin{array}{ccccc}
			\toprule \text { Method }  & \text {Frame1}\downarrow & \text {Frame2}\downarrow & \text{Frame3}\downarrow &\text {Frame4}\downarrow \\
			\midrule 
			\text { VPN~\cite{kalchbrenner2017video} } & \text{0.744} & \text{1.031} & \text{1.251} & \text{1.444} \\
			\text { ST-ResNet~\cite{zhang2017deep}} & \text{0.688} & \text{0.939} & \text{1.130} & \text{1.288} \\
			\text { FRNN~\cite{oliu2018folded} } & \text{0.682} & \text{0.823} & \text{0.989} & \text{1.183} \\
			\text { PredRNN~\cite{wang2017predrnn} } & \text{0.634} & \text{0.934} & \text{1.047} & \text{1.263} \\
			\text { PredRNN++~\cite{wang2018predrnn++}  } & \text{0.641} & \text{0.855} & \text{0.979} & \text{1.158} \\
			\text { E3d-LSTM~\cite{wang2018eidetic} } & \text{0.620} & \text{0.773} & \text{0.888} & \text{1.158} \\
			\text { MIM~\cite{wang2019memory} } & \text{0.554} & \text{0.737} & \text{0.887} & \text{0.999} \\
			\midrule
			\text{SwinLSTM} & \textbf{0.324} & \textbf{0.401} & \textbf{0.473} & \textbf{0.525} \\
			\bottomrule
		\end{array}
		$
	}
	\caption{  Quantitative results on the \textbf{TaxiBJ} dataset. We report the per-frame MSE of the 4 predictions.}
	\label{tab:taxi}
\end{table}

\subsection{Ablation Study}
\label{sec:Ablation}
In this section, we perform ablation studies on TaxiBJ and Human3.6m to analyze the impact of different elements on model performance. Specifically, we discuss three major elements: the reconstruction layer, patch size, and the number of Swin Transformer blocks.

\noindent{\bf The reconstruction layer.} The role of the reconstruction layer is to decode the spatiotemporal representations extracted by the SwinLSTM cell. We conduct experiments on transposed convolution, bilinear interpolation, and linear projection to analyze the impact of different decoding methods. Table~\ref{tab:reconstruction} shows that the transposed convolution performs much better than the other two methods.
\begin{table}[h]
	\centering
	\resizebox{\hsize}{!}{
		$
		\begin{array}{l|cccc}
			\toprule
			&\multicolumn{2}{c}{\text { TaxiBJ}} & \multicolumn{2}{c}{\text { Human3.6m}} \\
			\text { The reconstruction layer}& \text { MSE } & \text { SSIM } & \text { MSE  }/10 & \text { SSIM } \\
			\midrule 
			\text { Transposed convolution } &\textbf {0.390} & \textbf {0.980}  & \textbf {33.2} & \textbf {0.913} \\
			\text { Bilinear interpolation } &\text {0.794}  & \text {0.961} & \text {39.3}  & \text {0.887} \\
			\text { Linear Projection } & \text {0.415} & \text {0.979} & \text {72.5} & \text {0.773} \\
			\bottomrule
		\end{array}
		$
	}
	\caption{Ablation study on the three different methods of reconstruction layer with SwinLSTM on TaxiBJ and Human3.6m.}
	\label{tab:reconstruction}
\end{table}

\noindent{\bf Patch size.} The size of the image patch directly determines the length of the input token sequence. The smaller the size, the longer the length. To explore the effect of different patch sizes on prediction accuracy, we conduct experiments with patch sizes set to 2, 4, and 8 on TaxiBJ and Human3.6m. Table~\ref{tab:Patch} shows that the optimal patch size varies across datasets, which indicates that adjusting the patch size to an appropriate size is beneficial to improve the generalization of SwinLSTM.
\begin{table}[h]
	\centering
	$
	\begin{array}{c|cccc}
		\toprule
		&\multicolumn{2}{c}{\text { TaxiBJ}} & \multicolumn{2}{c}{\text { Human3.6m}} \\
		\text {Patch size}& \text { MSE } & \text { SSIM } & \text { MSE  }/10 & \text { SSIM } \\
		\midrule 
		\text { 2 } & \text {0.417} & \text {0.979} & \textbf{33.2}  & \textbf{0.913} \\
		\text { 4 } & \textbf{0.390} & \textbf{0.980} & \text {40.2} & \text {0.885} \\
		\text { 8 } & \text {0.393} & \text {0.980} & \text {51.9}  & \text {0.839} \\
		\bottomrule
	\end{array}
	$
	\caption{Ablation study on different patch sizes on TaxiBJ and Human3.6m.}
	\label{tab:Patch}
\end{table}

\begin{figure}[H]
	\centering
	\begin{tabular}{cc}
		\hspace{-0.8cm} \includegraphics[width=4.5cm]{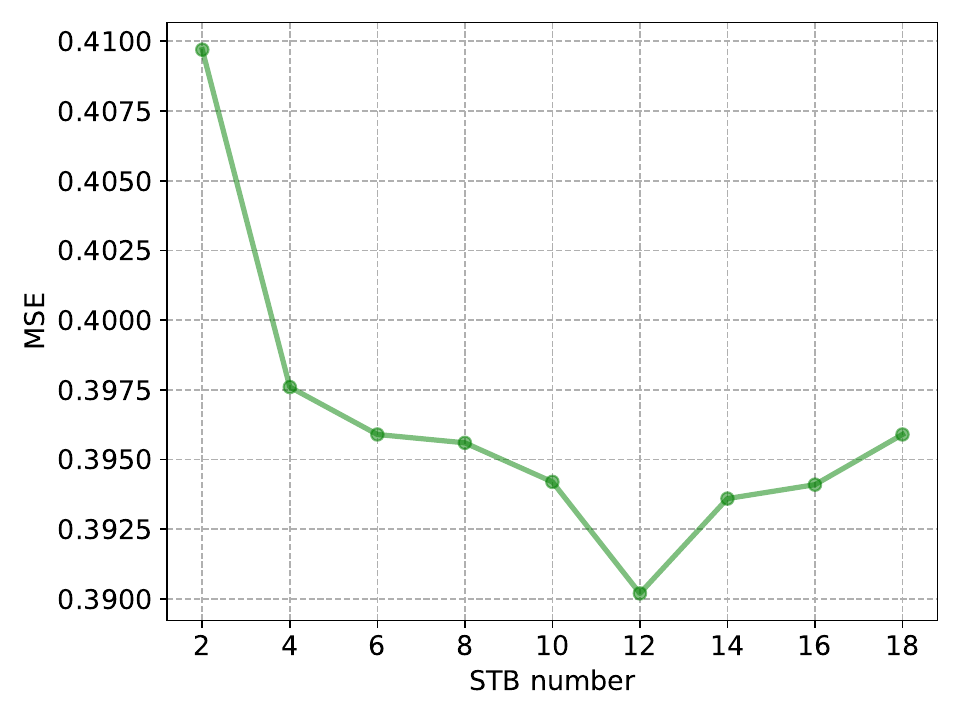}     &  \hspace{-0.5cm} \includegraphics[width=4.5cm]{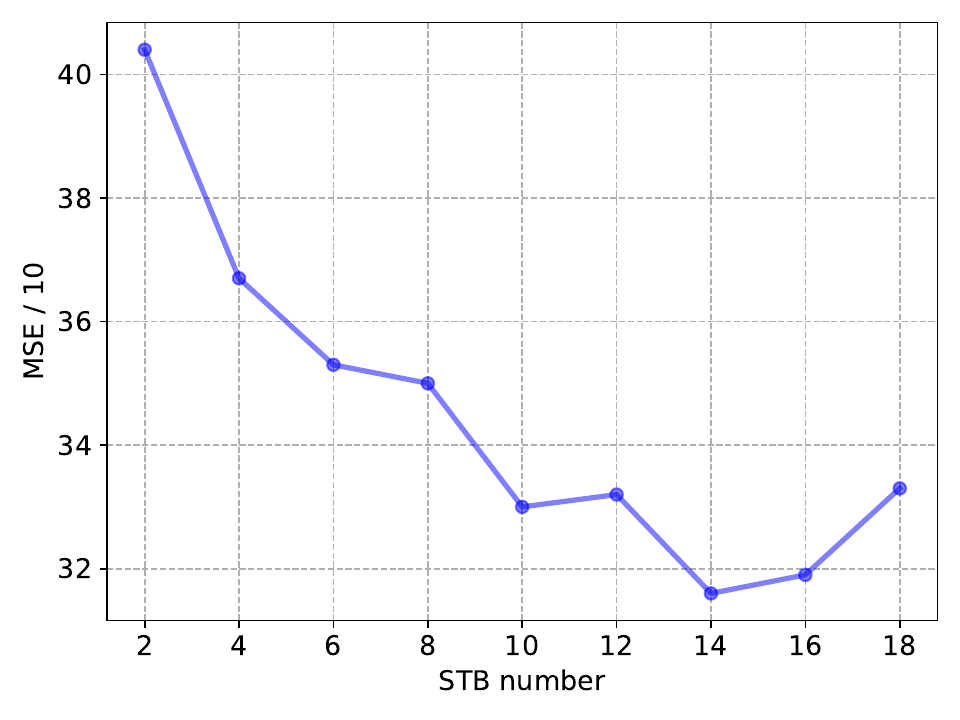} \\
		(a) TaxiBJ & (b) Human3.6m 
	\end{tabular} \\
	
	\caption{Ablation study on the different numbers of STB with SwinLSTM on TaxiBJ and Human3.6m.}
	\label{fig:line_chart}
\end{figure}

\noindent{\bf The number of Swin Transformer Blocks.} Based on the mechanism of Swin Transformer blocks (STB) modeling global spatial information, it is evident that the learning capacity of the model varies with the number of STB utilized. We explored the effect of the number of STB from 2 to 18 on model performance. Figure~\ref{fig:line_chart} present the results of different numbers of STB on TaxiBJ and Human3.6m.

\subsection{Feature map Visualization}
\label{Feature}
\begin{figure}
	\centering
	\resizebox{\hsize}{!}{
		\includegraphics{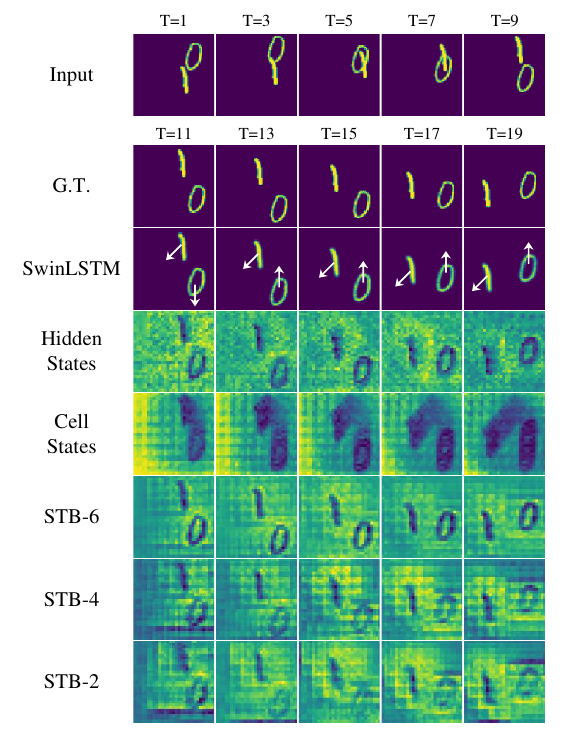}
	}
	\caption{Feature map visualization of a sample in the Moving MNIST test set. \textbf{Hidden States} and \textbf{Cell States} are the feature maps of  $\mathcal{H}_{t}$ and  $\mathcal{C}_{t}$ of the recurrent cell SwinLSTM (section~\ref{sec:SwinLSTM}) at different time steps. \textbf{STB-2},\textbf{STB-4}, and \textbf{STB-6} are the feature maps of the 2nd, 4th, and 6th layers of the Swin Transformer blocks (section~\ref{sec:STB}), respectively.}
	\label{fig:feature_map}
\end{figure}

To analyze the mechanism of the SwinLSTM cell (section~\ref{sec:SwinLSTM}), we randomly selected a sample from the Moving MNIST test set and visualized the feature maps during inference. Figure~\ref{fig:feature_map} shows the results, where white arrows have been added to the images in the third row to facilitate the observation of digit trajectories. The Hidden States capture the position of digits at the current time step, whereas the Cell States memorize the digit trajectory. Meanwhile, we conducted a visualization of the feature maps generated by the second, fourth, and sixth Swin Transformer blocks of the SwinLSTM-B. By observing the feature maps of STB-2 to STB-6, it is evident that STB gradually learns global spatial correlations as the number of interactions of window self-attention increases (section~\ref{sec:STB}). By analyzing the visualizations generated by our model, we aimed to gain valuable insights into its performance and behavior. These insights will inform future optimizations and improvements.

\section{Conclusion}
In this paper, we propose SwinLSTM, a new recurrent cell that incorporates Swin Transformer blocks and simplified LSTM, and constructs a predictive network for spatiotemporal prediction tasks based on it. Through extensive experimentation, it has been demonstrated that the proposed method, SwinLSTM, achieves excellent performance on various datasets, including Moving MNIST, TaxiBJ, Human3.6m, and KTH. These impressive results showcase the effectiveness and generalization ability of SwinLSTM. Notably, SwinLSTM outperforms its predecessor, ConvLSTM, in prediction accuracy, which suggests that learning global spatial dependencies allows the model to better capture spatiotemporal dependencies. Overall, this study highlights the potential of SwinLSTM as a promising approach for spatiotemporal modeling.

\section*{Acknowledgement}  This work was supported by the Natural Science Foundation Innovation Research Team Project of Hainan Province (320CXTD431), the National Key Research and Development Program of China (2021YFB1507104), the National Natural Science Foundation of China (32060413), and the Major Science and Technology Project of Hainan Province (ZDKJ2020013), which are gratefully acknowledged.

{\small
\bibliographystyle{ieee_fullname}
\bibliography{egbib}

\begin{thebibliography}{10}\itemsep=-1pt

\bibitem{bhattacharyya2018long}
Apratim Bhattacharyya, Mario Fritz, and Bernt Schiele.
\newblock Long-term on-board prediction of people in traffic scenes under
  uncertainty.
\newblock In {\em Proceedings of the IEEE Conference on Computer Vision and
  Pattern Recognition}, pages 4194--4202, 2018.

\bibitem{cao2023swin}
Hu Cao, Yueyue Wang, Joy Chen, Dongsheng Jiang, Xiaopeng Zhang, Qi Tian, and
  Manning Wang.
\newblock Swin-unet: Unet-like pure transformer for medical image segmentation.
\newblock In {\em Computer Vision--ECCV 2022 Workshops: Tel Aviv, Israel,
  October 23--27, 2022, Proceedings, Part III}, pages 205--218. Springer, 2023.

\bibitem{chang2021mau}
Zheng Chang, Xinfeng Zhang, Shanshe Wang, Siwei Ma, Yan Ye, Xiang Xinguang, and
  Wen Gao.
\newblock Mau: A motion-aware unit for video prediction and beyond.
\newblock {\em Advances in Neural Information Processing Systems},
  34:26950--26962, 2021.

\bibitem{chen20182}
Yunpeng Chen, Yannis Kalantidis, Jianshu Li, Shuicheng Yan, and Jiashi Feng.
\newblock A\^{} 2-nets: Double attention networks.
\newblock {\em Advances in neural information processing systems}, 31, 2018.

\bibitem{deng2009imagenet}
Jia Deng, Wei Dong, Richard Socher, Li-Jia Li, Kai Li, and Li Fei-Fei.
\newblock Imagenet: A large-scale hierarchical image database.
\newblock In {\em 2009 IEEE conference on computer vision and pattern
  recognition}, pages 248--255. Ieee, 2009.

\bibitem{dosovitskiy2020image}
Alexey Dosovitskiy, Lucas Beyer, Alexander Kolesnikov, Dirk Weissenborn,
  Xiaohua Zhai, Thomas Unterthiner, Mostafa Dehghani, Matthias Minderer, Georg
  Heigold, Sylvain Gelly, et~al.
\newblock An image is worth 16x16 words: Transformers for image recognition at
  scale.
\newblock {\em arXiv preprint arXiv:2010.11929}, 2020.

\bibitem{gao2019disentangling}
Hang Gao, Huazhe Xu, Qi-Zhi Cai, Ruth Wang, Fisher Yu, and Trevor Darrell.
\newblock Disentangling propagation and generation for video prediction.
\newblock In {\em Proceedings of the IEEE/CVF International Conference on
  Computer Vision}, pages 9006--9015, 2019.

\bibitem{guen2020disentangling}
Vincent~Le Guen and Nicolas Thome.
\newblock Disentangling physical dynamics from unknown factors for unsupervised
  video prediction.
\newblock In {\em Proceedings of the IEEE/CVF Conference on Computer Vision and
  Pattern Recognition}, pages 11474--11484, 2020.

\bibitem{hochreiter1997long}
Sepp Hochreiter and J{\"u}rgen Schmidhuber.
\newblock Long short-term memory.
\newblock {\em Neural computation}, 9(8):1735--1780, 1997.

\bibitem{ionescu2013human3}
Catalin Ionescu, Dragos Papava, Vlad Olaru, and Cristian Sminchisescu.
\newblock Human3. 6m: Large scale datasets and predictive methods for 3d human
  sensing in natural environments.
\newblock {\em IEEE transactions on pattern analysis and machine intelligence},
  36(7):1325--1339, 2013.

\bibitem{jia2016dynamic}
Xu Jia, Bert De~Brabandere, Tinne Tuytelaars, and Luc~V Gool.
\newblock Dynamic filter networks.
\newblock {\em Advances in neural information processing systems}, 29, 2016.

\bibitem{jin2020exploring}
Beibei Jin, Yu Hu, Qiankun Tang, Jingyu Niu, Zhiping Shi, Yinhe Han, and
  Xiaowei Li.
\newblock Exploring spatial-temporal multi-frequency analysis for high-fidelity
  and temporal-consistency video prediction.
\newblock In {\em Proceedings of the IEEE/CVF Conference on Computer Vision and
  Pattern Recognition}, pages 4554--4563, 2020.

\bibitem{jin2018varnet}
Beibei Jin, Yu Hu, Yiming Zeng, Qiankun Tang, Shice Liu, and Jing Ye.
\newblock Varnet: Exploring variations for unsupervised video prediction.
\newblock In {\em 2018 IEEE/RSJ International Conference on Intelligent Robots
  and Systems (IROS)}, pages 5801--5806. IEEE, 2018.

\bibitem{kalchbrenner2017video}
Nal Kalchbrenner, A{\"a}ron Oord, Karen Simonyan, Ivo Danihelka, Oriol Vinyals,
  Alex Graves, and Koray Kavukcuoglu.
\newblock Video pixel networks.
\newblock In {\em International Conference on Machine Learning}, pages
  1771--1779. PMLR, 2017.

\bibitem{kingma2015adam}
Diederik~P Kingma and Jimmy Ba.
\newblock Adam: A method for stochastic optimization.
\newblock In {\em ICLR (Poster)}, 2015.

\bibitem{kwon2019predicting}
Yong-Hoon Kwon and Min-Gyu Park.
\newblock Predicting future frames using retrospective cycle gan.
\newblock In {\em Proceedings of the IEEE/CVF Conference on Computer Vision and
  Pattern Recognition}, pages 1811--1820, 2019.

\bibitem{lecun1998mnist}
Yann LeCun.
\newblock The mnist database of handwritten digits.
\newblock {\em http://yann. lecun. com/exdb/mnist/}, 1998.

\bibitem{lee2018stochastic}
Alex~X Lee, Richard Zhang, Frederik Ebert, Pieter Abbeel, Chelsea Finn, and
  Sergey Levine.
\newblock Stochastic adversarial video prediction.
\newblock {\em arXiv preprint arXiv:1804.01523}, 2018.

\bibitem{lee2021video}
Sangmin Lee, Hak~Gu Kim, Dae~Hwi Choi, Hyung-Il Kim, and Yong~Man Ro.
\newblock Video prediction recalling long-term motion context via memory
  alignment learning.
\newblock In {\em Proceedings of the IEEE/CVF Conference on Computer Vision and
  Pattern Recognition}, pages 3054--3063, 2021.

\bibitem{liu2021swin}
Ze Liu, Yutong Lin, Yue Cao, Han Hu, Yixuan Wei, Zheng Zhang, Stephen Lin, and
  Baining Guo.
\newblock Swin transformer: Hierarchical vision transformer using shifted
  windows.
\newblock In {\em Proceedings of the IEEE/CVF International Conference on
  Computer Vision}, pages 10012--10022, 2021.

\bibitem{luo2016understanding}
Wenjie Luo, Yujia Li, Raquel Urtasun, and Richard Zemel.
\newblock Understanding the effective receptive field in deep convolutional
  neural networks.
\newblock {\em Advances in neural information processing systems}, 29, 2016.

\bibitem{oliu2018folded}
Marc Oliu, Javier Selva, and Sergio Escalera.
\newblock Folded recurrent neural networks for future video prediction.
\newblock In {\em Proceedings of the European Conference on Computer Vision
  (ECCV)}, pages 716--731, 2018.

\bibitem{schuldt2004recognizing}
Christian Schuldt, Ivan Laptev, and Barbara Caputo.
\newblock Recognizing human actions: a local svm approach.
\newblock In {\em Proceedings of the 17th International Conference on Pattern
  Recognition, 2004. ICPR 2004.}, volume~3, pages 32--36. IEEE, 2004.

\bibitem{shi2015convolutional}
Xingjian Shi, Zhourong Chen, Hao Wang, Dit-Yan Yeung, Wai-Kin Wong, and
  Wang-chun Woo.
\newblock Convolutional lstm network: A machine learning approach for
  precipitation nowcasting.
\newblock {\em Advances in neural information processing systems}, 28, 2015.

\bibitem{shi2017deep}
Xingjian Shi, Zhihan Gao, Leonard Lausen, Hao Wang, Dit-Yan Yeung, Wai-kin
  Wong, and Wang-chun Woo.
\newblock Deep learning for precipitation nowcasting: A benchmark and a new
  model.
\newblock {\em Advances in neural information processing systems}, 30, 2017.

\bibitem{srivastava2015unsupervised}
Nitish Srivastava, Elman Mansimov, and Ruslan Salakhudinov.
\newblock Unsupervised learning of video representations using lstms.
\newblock In {\em International conference on machine learning}, pages
  843--852. PMLR, 2015.

\bibitem{touvron2021training}
Hugo Touvron, Matthieu Cord, Matthijs Douze, Francisco Massa, Alexandre
  Sablayrolles, and Herv{\'e} J{\'e}gou.
\newblock Training data-efficient image transformers \& distillation through
  attention.
\newblock In {\em International Conference on Machine Learning}, pages
  10347--10357. PMLR, 2021.

\bibitem{vaswani2021scaling}
Ashish Vaswani, Prajit Ramachandran, Aravind Srinivas, Niki Parmar, Blake
  Hechtman, and Jonathon Shlens.
\newblock Scaling local self-attention for parameter efficient visual
  backbones.
\newblock In {\em Proceedings of the IEEE/CVF Conference on Computer Vision and
  Pattern Recognition}, pages 12894--12904, 2021.

\bibitem{vaswani2017attention}
Ashish Vaswani, Noam Shazeer, Niki Parmar, Jakob Uszkoreit, Llion Jones,
  Aidan~N Gomez, {\L}ukasz Kaiser, and Illia Polosukhin.
\newblock Attention is all you need.
\newblock {\em Advances in neural information processing systems}, 30, 2017.

\bibitem{villegas2017decomposing}
Ruben Villegas, Jimei Yang, Seunghoon Hong, Xunyu Lin, and Honglak Lee.
\newblock Decomposing motion and content for natural video sequence prediction.
\newblock {\em arXiv preprint arXiv:1706.08033}, 2017.

\bibitem{wang2018predrnn++}
Yunbo Wang, Zhifeng Gao, Mingsheng Long, Jianmin Wang, and S~Yu Philip.
\newblock Predrnn++: Towards a resolution of the deep-in-time dilemma in
  spatiotemporal predictive learning.
\newblock In {\em International Conference on Machine Learning}, pages
  5123--5132. PMLR, 2018.

\bibitem{wang2018eidetic}
Yunbo Wang, Lu Jiang, Ming-Hsuan Yang, Li-Jia Li, Mingsheng Long, and Li
  Fei-Fei.
\newblock Eidetic 3d lstm: A model for video prediction and beyond.
\newblock In {\em International conference on learning representations}, 2018.

\bibitem{wang2017predrnn}
Yunbo Wang, Mingsheng Long, Jianmin Wang, Zhifeng Gao, and Philip~S Yu.
\newblock Predrnn: Recurrent neural networks for predictive learning using
  spatiotemporal lstms.
\newblock {\em Advances in neural information processing systems}, 30, 2017.

\bibitem{wang2019memory}
Yunbo Wang, Jianjin Zhang, Hongyu Zhu, Mingsheng Long, Jianmin Wang, and
  Philip~S Yu.
\newblock Memory in memory: A predictive neural network for learning
  higher-order non-stationarity from spatiotemporal dynamics.
\newblock In {\em Proceedings of the IEEE/CVF Conference on Computer Vision and
  Pattern Recognition}, pages 9154--9162, 2019.

\bibitem{wang2004image}
Zhou Wang, Alan~C Bovik, Hamid~R Sheikh, and Eero~P Simoncelli.
\newblock Image quality assessment: from error visibility to structural
  similarity.
\newblock {\em IEEE transactions on image processing}, 13(4):600--612, 2004.

\bibitem{wu2022p2t}
Yu-Huan Wu, Yun Liu, Xin Zhan, and Ming-Ming Cheng.
\newblock P2t: Pyramid pooling transformer for scene understanding.
\newblock {\em IEEE Transactions on Pattern Analysis and Machine Intelligence},
  2022.

\bibitem{xie2021segformer}
Enze Xie, Wenhai Wang, Zhiding Yu, Anima Anandkumar, Jose~M Alvarez, and Ping
  Luo.
\newblock Segformer: Simple and efficient design for semantic segmentation with
  transformers.
\newblock {\em Advances in Neural Information Processing Systems},
  34:12077--12090, 2021.

\bibitem{xu2018predcnn}
Ziru Xu, Yunbo Wang, Mingsheng Long, Jianmin Wang, and M KLiss.
\newblock Predcnn: Predictive learning with cascade convolutions.
\newblock In {\em IJCAI}, pages 2940--2947, 2018.

\bibitem{yu2020efficient}
Wei Yu, Yichao Lu, Steve Easterbrook, and Sanja Fidler.
\newblock Efficient and information-preserving future frame prediction and
  beyond.
\newblock 2020.

\bibitem{zhang2017deep}
Junbo Zhang, Yu Zheng, and Dekang Qi.
\newblock Deep spatio-temporal residual networks for citywide crowd flows
  prediction.
\newblock In {\em Thirty-first AAAI conference on artificial intelligence},
  2017.

\bibitem{zhang2021multi}
Pengchuan Zhang, Xiyang Dai, Jianwei Yang, Bin Xiao, Lu Yuan, Lei Zhang, and
  Jianfeng Gao.
\newblock Multi-scale vision longformer: A new vision transformer for
  high-resolution image encoding.
\newblock In {\em Proceedings of the IEEE/CVF International Conference on
  Computer Vision}, pages 2998--3008, 2021.

\bibitem{zhao2021point}
Hengshuang Zhao, Li Jiang, Jiaya Jia, Philip~HS Torr, and Vladlen Koltun.
\newblock Point transformer.
\newblock In {\em Proceedings of the IEEE/CVF International Conference on
  Computer Vision}, pages 16259--16268, 2021.

\bibitem{zheng2021rethinking}
Sixiao Zheng, Jiachen Lu, Hengshuang Zhao, Xiatian Zhu, Zekun Luo, Yabiao Wang,
  Yanwei Fu, Jianfeng Feng, Tao Xiang, Philip~HS Torr, et~al.
\newblock Rethinking semantic segmentation from a sequence-to-sequence
  perspective with transformers.
\newblock In {\em Proceedings of the IEEE/CVF conference on computer vision and
  pattern recognition}, pages 6881--6890, 2021.

\end{thebibliography}
}
\end{document}